\documentclass[10pt,twocolumn,letterpaper]{article}

\usepackage[pagenumbers]{cvpr} 

\usepackage{tabularx,booktabs,array}
\usepackage{multirow}
\usepackage{amsmath}
\usepackage{amsfonts}
\usepackage{pifont}
\definecolor{cvprblue}{rgb}{0.21,0.49,0.74}
\usepackage[pagebackref,breaklinks,colorlinks,allcolors=cvprblue]{hyperref}

\def\paperID{6511} 
\def\confName{CVPR}
\def\confYear{2026}

\title{Egocentric Visibility-Aware Human Pose Estimation}

\author{Peng Dai, Yu Zhang, Yiqiang Feng, Zhen Fan, Yang Zhang \\
PICO \\
{\tt\small \{daipeng.2022, zhangyu.118, fengyiqiang, fanzhen.0315, zhangyang.0621\}@bytedance.com}
}

\begin{document}
\maketitle
\begin{abstract}
Egocentric human pose estimation (HPE) using a head-mounted device is crucial for various VR and AR applications, but it faces significant challenges due to keypoint invisibility. Nevertheless, none of the existing egocentric HPE datasets provide keypoint visibility annotations, and the existing methods often overlook the invisibility problem, treating visible and invisible keypoints indiscriminately during estimation. As a result, their capacity to accurately predict visible keypoints is compromised. In this paper, we first present Eva-3M, a large-scale egocentric visibility-aware HPE dataset comprising over 3.0M frames, with 435K of them annotated with keypoint visibility labels. Additionally, we augment the existing EMHI dataset with keypoint visibility annotations to further facilitate the research in this direction. Furthermore, we propose EvaPose, a novel egocentric visibility-aware HPE method that explicitly incorporates visibility information to enhance pose estimation accuracy. Extensive experiments validate the significant value of ground-truth visibility labels in egocentric HPE settings, and demonstrate that our EvaPose achieves state-of-the-art performance in both Eva-3M and EMHI datasets.
\end{abstract}    
\begin{figure*}[htp]
  \centering
  \includegraphics[width=0.96\linewidth]{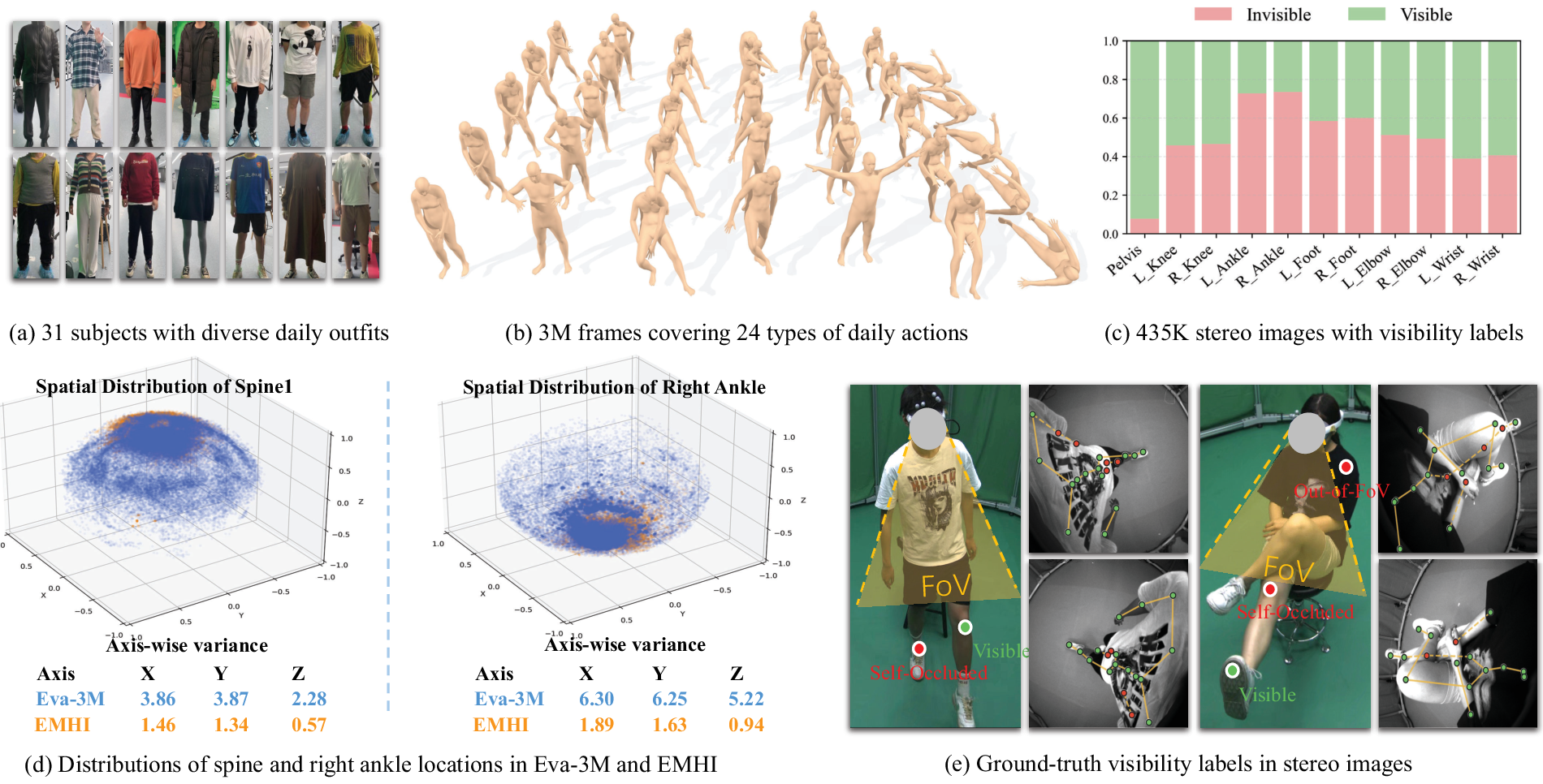}
  \caption{We introduce Eva-3M, a large-scale egocentric visibility-aware dataset comprising over 3.0M frames from (a) 31 subjects in daily outfits (b) performing 24 types of daily actions, and of which (c) 435K are annotated with keypoint visibility labels. (d) shows the normalized spatial distribution comparison between Eva-3M and EMHI, indicating that Eva-3M has a wider range of motion diversity than EMHI. (e) illustrates a few representative examples of keypoint invisibility due to self-occlusion and out-of-FoV in Eva-3M dataset.}
  \label{fig:dataset}
\end{figure*}

\section{Introduction}
\label{sec:intro}
Egocentric human pose estimation (HPE) using a head-mounted device (HMD) has gained great attention in recent years for its numerous applications in virtual reality (VR), augmented reality (AR), robotic control, etc~\cite{azam2024survey}. Different from outside-in HPE with external cameras, a key challenge of egocentric HPE lies in the keypoint invisibility problem~\cite{yang2024egoposeformer, lee2025rewind, camiletto2025frame}, which usually results from two causes. As shown in Fig.~\ref{fig:dataset} (e), the first cause arises from severe self-occlusions of body parts, particularly the lower body. Another cause is the limited field-of-view (FoV) of head-mounted cameras, making it difficult to fully capture the body especially when hands and legs are stretched out.

Annotating the visibility of each keypoint is a labor-intensive and time-consuming process. To the best of our knowledge, none of the existing egocentric HPE datasets include visibility annotations for keypoints. Consequently, current egocentric HPE methods~\cite{wang2023scene, yang2024egoposeformer, fan2025emhi} often neglect the invisibility issue and treat visible and invisible keypoints indiscriminately in pose estimation. For invisible keypoints, the absence of direct visual evidence introduces inherent ambiguity in predicting their 3D positions. As our experiments demonstrate, this undifferentiated treatment compromises the estimation accuracy of visible keypoints.

We introduce Eva-3M, a large-scale, visibility-aware egocentric HPE dataset comprising over 3.0M frames, including 435K with detailed keypoint visibility labels. As shown in Fig.~\ref{fig:dataset}, Eva-3M was collected from 31 subjects performing 24 types of daily activities using a commercial Pico4 Ultra VR-MR headset, offering realistic insights into HPE with real VR devices. It stands as the first and largest real egocentric dataset to provide both ground-truth SMPL~\cite{SMPL:2015} poses and visibility labels. To foster broader research, we also extend our contribution by providing keypoint visibility labels for the existing EMHI~\cite{fan2025emhi} dataset. 

To effectively leverage keypoint visibility and mitigate the interference of invisible keypoints on visible ones, we propose EvaPose, a novel visibility-aware framework for egocentric HPE. As depicted in Fig.~\ref{fig:overview}, EvaPose comprises three key components: (1) To ensure the generation of plausible 3D poses particularly for invisible keypoints, we employ a Vector Quantized-VAE (VQ-VAE)~\cite{van2017neural} model pre-trained on extensive motion capture (mocap) datasets to embed realistic human pose priors. (2) EvaPose presents a visibility-aware 3D estimation network to predict per-frame 3D keypoints and their visibility states for the first time. This is complemented by a visibility-based loss-weighting scheme that applies distinct supervision to visible and invisible keypoints during training. (3) An iterative intra-and inter-frame attention module is designed to refine the initial predictions. Within this module, per-frame 3D keypoints and their visibility scores iteratively interact with stereo image features and with each other across the whole sequence, achieving robust multi-view and temporal fusion. Finally, the fused features are processed by the pre-trained VQ-VAE decoder to reconstruct the final, high-fidelity 3D poses.


In summary, our main contributions are as follows:
\begin{itemize}
\item \textbf{Eva-3M Dataset}: A large-scale, real egocentric HPE dataset with keypoint visibility labels for the first time.
\item \textbf{EMHI Enrichment}: We augment the existing EMHI dataset~\cite{fan2025emhi} by contributing keypoint visibility labels.
\item \textbf{EvaPose Method}: A novel visibility-aware framework that explicitly utilizes keypoint visibility information to significantly enhance the 3D pose estimation accuracy.
\item \textbf{Superior Performance}: Extensive experiments validate the superiority of our method over previous methods.
\end{itemize}
\section{Related Work}
\label{sec:related_work}
\subsection{Egocentric HPE Datasets}
Recent years have seen the release of large-scale datasets for egocentric video understanding~\cite{ye2024mm}, egocentric action recognition~\cite{grauman2022ego4d, grauman2024ego}, and egocentric life assistant~\cite{yang2025egolife}.
However, in the field of egocentric HPE, the availability of large-scale datasets still remains limited. Collecting a real-world egocentric HPE dataset is challenging due to specialized devices and complex setups are required for accurate markerless motion capture. Consequently, prior work has predominantly relied on synthetic datasets~\cite{xu2019mo, tome2019xr, akada2022unrealego, wang2023scene, akada20243d, wang2024egocentric, cuevas2024simpleego}. 
Although synthetic data enables scalable generation with perfect ground-truth annotations, the inherent domain gap between synthetic and real-captured data inevitably limits the performance of the models when deployed in real-world scenarios.


For real-world egocentric HPE datasets, most existing works use custom-made wearable capture rigs, such as EgoCap~\cite{rhodin2016egocap}, EgoPW-Scene~\cite{wang2023scene}, UE-RW~\cite{akada20243d}, Ego4View-RW~\cite{hakada2025egorear}, and SELF~\cite{camiletto2025frame} dataset. However, these capture rigs differ significantly from commercial VR/AR devices in both design and imaging characteristics.
For instance, custom setups often employ protruding camera placements to minimize self-occlusions, while commercial VR/AR devices follow a slim and lightweight design, which typically induces more severe self-occlusions. In response to this discrepancy, several recent datasets have been constructed using actual VR/AR devices. Nymeria~\cite{ma2024nymeria} uses Project Aria glasses~\cite{somasundaram2023project} to collect 300 hours of in-the-wild daily motion data. EgoBody3M~\cite{zhao2024egobody3m} introduces the first egocentric HPE dataset captured with a realistic VR headset configuration, while EMHI~\cite{fan2025emhi} provides a multi-modal egocentric human motion dataset combining real VR devices with body-worn inertial measurement units (IMUs). Despite these advances, neither EgoBody3M~\cite{zhao2024egobody3m} nor EMHI~\cite{fan2025emhi} addresses the issue of keypoint invisibility, and neither dataset provides visibility annotations for keypoints. 

\subsection{Egocentric HPE Methods}
Existing methods for estimating the wearer’s body pose vary by input modality, including sparse IMUs~\cite{aliakbarian2023hmd, AvatarPoser, zheng2023realistic, dai2024hmd, xia2025envposer}, front-facing egocentric cameras~\cite{li2023ego, yi2025estimating, guzov2025hmd, hong2025egolm}, and down-facing egocentric cameras~\cite{xu2019mo, tome2019xr, tome2020selfpose, zhao2021egoglass, akada2022unrealego, akada20243d, kang2024attention, yang2024egoposeformer, zhao2024egobody3m, fan2025emhi, lee2025rewind, camiletto2025frame}. Among these, front-facing camera approaches often treat pose estimation as a motion generation or inpainting task, since the wearer’s body is largely out of view. In this work, we focus primarily on methods using down-facing egocentric cameras, which are most relevant to our setting.

Early efforts in this line of work mainly addressed 3D pose estimation from visual-only inputs. EgoGlass~\cite{zhao2021egoglass} first predicts 2D heatmaps from each view and then applies a 3D lifting module to reconstruct 3D keypoints from multi-view 2D heatmaps. UnrealEgo~\cite{akada2022unrealego} follows a similar pipeline but introduces a dual weight-sharing encoder architecture to better leverage stereo information for heatmap estimation. EgoPoseFormer~\cite{yang2024egoposeformer} further advances this direction with a Deformable Self-Attention (DSA) mechanism for effective multi-view feature fusion, and adopts a DETR-style~\cite{carion2020end} Transformer for coarse-to-fine pose estimation. 

More recent studies incorporate camera or device poses from SLAM systems to improve both 3D pose estimation accuracy and physical plausibility. FRAME~\cite{camiletto2025frame} first estimates per-frame 3D poses in local coordinates, aligns them into a global coordinate system using device poses, and refines the sequence of 3D poses with a temporal fusion module. REWIND~\cite{lee2025rewind} proposes a diffusion-based approach~\cite{ho2020denoising} that generates global whole-body motion conditioned on stereo egocentric images and camera poses.


Despite these advances, existing egocentric HPE methods overlook the adverse effect of invisible keypoints on the estimation accuracy of visible ones. In contrast, we propose a visibility-aware 3D pose estimation network along with a visibility-based loss-weighting scheme, which explicitly differentiates between visible and invisible keypoints during training to mitigate such interference.

\section{Eva-3M Dataset}
\label{sec:dataset}

The primary challenge in egocentric HPE lies in the prevalent issue of keypoint invisibility. As shown in Fig.~\ref{fig:dataset} (a), which illustrates the visibility statistics of 11 selected keypoints in our Eva-3M dataset, limb keypoints are invisible nearly half of the time. Despite its significant impact, none of the existing datasets provide visibility annotations. To address this gap, we introduce Eva-3M, the first and largest real-world egocentric dataset offering both ground-truth SMPL poses and comprehensive visibility labels, with substantially greater motion diversity. This combination supports more reliable training and evaluation of egocentric pose estimation models under realistic conditions.

A detailed comparison in Tab.~\ref{tab:dataset_comparison} highlights the advantages of our Eva-3M across several dimensions. First, unlike prior datasets built with custom capture rigs, Eva-3M is captured using a commercial Pico4 Ultra VR-MR headset, providing direct insights into HPE performance in real VR devices. Second, as depicted in Fig.~\ref{fig:dataset}(b) and (c), Eva-3M includes 3.0 million synchronized frames from 31 subjects performing 24 categories of common VR motions, offering richer motion diversities than existing benchmarks such as EMHI~\cite{fan2025emhi}. To quantify this, we compare the motion diversity between Eva-3M and EMHI by randomly sampling 70K root-relative 3D keypoint coordinates from both. For better visualization, the root-relative coordinates are normalized by their maximum distance. The spatial distributions of an internal keypoint (spine1) and an extremity keypoint (right ankle) are visualized in Fig.~\ref{fig:dataset}(d), where Eva-3M covers a noticeably broader spatial range. We further compute the variance of the original (non-normalized) 3D coordinates in decimeters, and observe consistently larger variances for both keypoints in Eva-3M, confirming its wider range of motion diversities. Third, Eva-3M provides the first large-scale keypoint visibility annotations across 435K frames. Representative examples of these annotations are visualized in Fig.~\ref{fig:dataset}(e).

Eva-3M consists of 1,353 motion sequences recorded at 30 fps. Each frame includes paired stereo grayscale images (640$\times$480), ground-truth SMPL parameters in both camera and world coordinates, and corresponding ground-truth 2D/3D keypoints. Further details regarding the data capture system, ground-truth acquisition, and visibility annotation process are provided in the supplementary materials.

\begin{table}[tp]
  \small
  \centering
  {\setlength{\tabcolsep}{6pt}
   \begin{tabularx}{\columnwidth}{
     >{\raggedright\arraybackslash}p{2.2cm} 
     *{7}{c@{\hspace{4pt}}}                     
   }
     \toprule
     Dataset & R/S & Cams & Frames & Vis & SMPL & Subj & Act \\
     \midrule
     Mo$^2$Cap$^2$~\cite{xu2019mo}            & Syn.  & 1 & 530K  & $\times$   & $\times$   & 700& 3K \\
     xR-EgoPose~\cite{tome2019xr}            & Syn.  & 1 & 383K  & $\times$   & $\times$   & 46  & -  \\
     UnrealEgo~\cite{akada2022unrealego}& Syn.  & 2 & 450K  & $\times$   & $\times$   & 17 & 30 \\
     EgoGTA~\cite{wang2023scene}      & Syn.  & 1 & 320K  & $\times$   & $\times$   & 5  & -  \\
     SynthEgo~\cite{cuevas2024simpleego}      & Syn.  & 2 & 60K  & $\times$   & \checkmark   & 6K  & -  \\    
     EgoCap~\cite{rhodin2016egocap}     & Real & 2 & 75K   & $\times$   & $\times$   & 8  & -  \\
     EgoGlobal~\cite{wang2021estimating} & Real & 1 & 12K   & $\times$   & $\times$   & 2  & 13 \\
     EgoGlass~\cite{zhao2021egoglass}   & Real  & 2 & 173K  & $\times$   & $\times$   & 10 & -  \\
     UE-RW~\cite{akada20243d}           & Real  & 2 & 260K  & $\times$   & $\times$   & 16 & -  \\
     SELF~\cite{camiletto2025frame}     & Real  & 2 & 1.6M  & $\times$   & $\times$   & 14 & -  \\
     EMHI~\cite{fan2025emhi}     & Real  & 2 & 3.1M  & $\times$   & \checkmark   & 58 & 39  \\     
     EgoBody3M~\cite{zhao2024egobody3m}     & Real  & 4 & 3.4M  & $\times$   & $\times$   & 120 & 30  \\    
     \midrule
     \textbf{Eva-3M}                     & Real  & 2 & 3.0M & \checkmark & \checkmark & 31 & 24 \\
     \bottomrule
   \end{tabularx}
  }
  \caption{Comparison of egocentric motion datasets. R/S denotes whether the dataset is collected in real or synthetic setting. Cams indicates the number of egocentric cameras. Vis denotes the availability of keypoint visibility labels. Subj is the number of subjects, and Act is the number of action categories.}
  \label{tab:dataset_comparison}
\end{table}

\begin{figure*}[htp]
  \centering
  \includegraphics[width=1.0\linewidth]{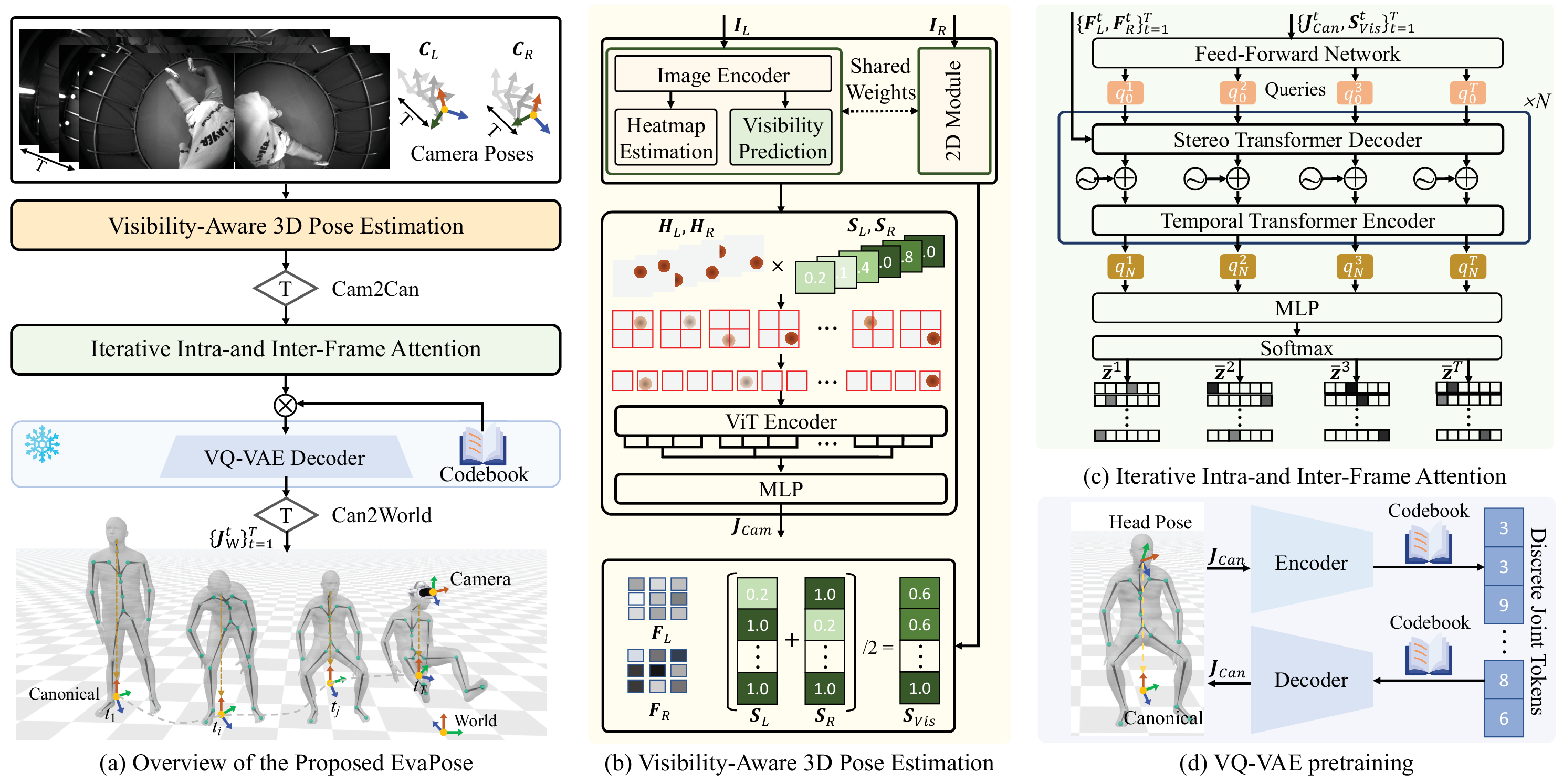}
  \caption{Overview of the proposed EvaPose. Given a sequence of egocentric observations, we first propose a visibility-aware 3D pose estimation network to extract stereo image features, and predict per-frame 3D keypoints in the camera coordinate system (defined as the left camera coordinate system in this paper) and their corresponding visibility confidence scores. Then, the predicted 3D keypoints are transformed to the canonical coordinate system with the help of camera poses from SLAM system. Next, an iterative intra-and inter-frame attention network is used for temporal feature fusion. Finally, we estimate the 3D poses with the pre-trained VQ-VAE decoder.}
  \label{fig:overview}
\end{figure*}

\section{EvaPose}
\label{sec:method}

\subsection{Overview}
\label{subsec:overview}
Our task is to estimate the body poses of the wearer using egocentric sensors from a head-mounted device (HMD). Following recent works~\cite{camiletto2025frame, lee2025rewind}, our method takes as input both stereo egocentric videos and camera poses from the HMD’s built-in SLAM system. Formally, our proposed EvaPose models the human pose estimation problem conditioned on a sequence of egocentric observations over a window of $T$ frames:
\begin{equation}
f_{\phi}(\boldsymbol{J}_{W}^{1:T}\,|\,  \boldsymbol{I}_{L}^{1:T}, \boldsymbol{I}_{R}^{1:T},
\boldsymbol{C}_{L}^{1:T},
\boldsymbol{C}_{R}^{1:T})
\label{eq:network_modelling}
\end{equation}
where $\phi$ denotes the learnable network parameters, and $\boldsymbol{J}_{W}^{t}$ represents the set of $N_J$ SMPL~\cite{loper2023smpl} keypoints in the world coordinate system at frame $t$. The inputs consist of stereo egocentric images
$\boldsymbol{I}_{v\in\{L, R\}}^{t} \in \mathbb{R} ^{C\times H\times W}$ from the left ($L$) and right ($R$) views, along with their corresponding camera poses $\boldsymbol{C}_{v\in\{L, R\}}^t=[\boldsymbol{R}_{v}^t|\boldsymbol{T}_{v}^t] \in \mathbb{R} ^{3\times 4}$, which include camera rotation $\boldsymbol{R}_{v}^t \in \mathbb{R} ^{3\times 3}$ and translation $\boldsymbol{T}_{v}^t \in \mathbb{R} ^{3\times 1}$. The camera poses are provided by the HMD's SLAM system.


As illustrated in Fig.~\ref{fig:overview}, EvaPose mainly consists of three components. 
(1) A VQ-VAE~\cite{van2017neural} pretrained on extensive mocap datasets encodes plausible human poses into a discrete codebook, serving as a strong pose prior (Sec.~\ref{subsec:pose_prior}).
(2) A visibility-aware 3D pose estimation network extracts stereo image features $[\boldsymbol{F}_L^{t}, \boldsymbol{F}_R^{t}]$, and jointly predicts per-frame 3D keypoints $\boldsymbol{J}_{Cam}^{t}$ along with their visibility confidence scores $\boldsymbol{S}_{Vis}^{t}$ (Sec.~\ref{subsec:visibility_aware_3d_pose}).
(3) An iterative intra-and inter-frame attention network fuses the above data over the whole time window for 3D pose refinement (Sec.~\ref{subsec:iterative_3d_refinement}).

\subsection{Learning the Pose Prior}
\label{subsec:pose_prior}
This section describes how we encode 3D human poses into a discrete representation using VQ-VAE~\cite{van2017neural}. Previous methods typically adopt either the HumanML3D~\cite{guo2022generating} representation or local pose parameters~\cite{dwivedi2024tokenhmr} as the pose representation. However, both approaches overlook shape parameters and present limitations in our context. Since our goal is to estimate accurate 3D keypoint positions for different users with varying body proportions and heights, we represent human poses using 3D keypoints defined in a canonical coordinate system. As illustrated in Fig.~\ref{fig:overview} (d), this coordinate system is constructed by projecting the head joint onto the floor plane, aligning the vertical ($y$) axis accordingly, and using the projection of the head’s horizontal orientation as the horizontal axis. A key advantage of this representation is its invariance to arbitrary translations in the floor plane and rotations around the vertical axis.





The overall VQ-VAE architecture, illustrated in Fig.~\ref{fig:overview}(d), consists of an encoder $\boldsymbol{E}$, a decoder $\boldsymbol{D}$, and a learnable codebook $\boldsymbol{CB}=\{\boldsymbol{c}_k\}_{k=1}^K$, where each code $\boldsymbol{c}_k \in \mathbb{R} ^ {D}$ and $D$ is the code dimension. Given 3D keypoints $\boldsymbol{J}_{Can}$ in the canonical coordinate system as input, the encoder produces a latent feature sequence $\boldsymbol{z}=\boldsymbol{E}(\boldsymbol{J}_{Can})=[\boldsymbol{z}_1,\boldsymbol{z}_2,...,\boldsymbol{z}_M]$, where each $\boldsymbol{z}_i \in \mathbb{R} ^ {D}$ and $M$ is the number of features. Each $\boldsymbol{z}_i$ is quantized to tokens by finding the most similar code element in the codebook. During training, we adopt the strategy from prior work~\cite{dwivedi2024tokenhmr}, utilizing exponential moving average updates and periodic codebook resets to prevent codebook collapse. Meanwhile, a combination of three loss terms including the reconstruction loss, the embedding loss, and the commitment loss are used to optimize the network.

\subsection{Visibility-Aware 3D Pose Estimation}
\label{subsec:visibility_aware_3d_pose}
In contrast to prior egocentric HPE methods~\cite{wang2024egocentric, yang2024egoposeformer, camiletto2025frame}, we introduce a visibility-aware 3D pose estimation network which, for the first time, explicitly predicts the visibility state of each keypoint. By leveraging these predicted states, EvaPose differentiates between visible and invisible keypoints during training, leading to improved HPE accuracy. For notational simplicity, we omit frame index $t$ in this subsection, as all operations are applied identically per frame.

As illustrated in Fig.~\ref{fig:overview} (b), given a stereo image pair $\boldsymbol{I}_{v\in\{L, R\}}  \in \mathbb{R} ^{C\times H\times W}$ as input, an image encoder first extracts visual features $\boldsymbol{F}_{L}$ and $\boldsymbol{F}_{R}$. These features are then processed by two separate lightweight decoders: one estimates 2D heatmaps $\boldsymbol{H}_{v}  \in \mathbb{R} ^{N_J\times H'\times W'}$, and the other predicts visibility scores $\boldsymbol{S}_{v} \in \mathbb{R} ^ {N_J}$, where $H'$ and $W'$ denote the spatial dimensions of the heatmaps. Following the design of ViTPose~\cite{xu2022vitpose}, the heatmap decoder comprises a series of deconvolution layers followed by a final prediction layer. The visibility decoder, for simplicity, is implemented as a convolution layer followed by a Multi-Layer Perceptron (MLP). Subsequently, a ViT encoder integrates the multi-view heatmaps and visibility scores for 3D pose estimation. Let $\{\boldsymbol{H}_{i,v}\}$ and $\{s_{i,v}\}$ denote the sets of $2N_J$ joint heatmaps and visibility scores, respectively. We first compute visibility-aware heatmaps as $\boldsymbol{H}'_{i,v}=s_{i,v}\cdot \boldsymbol{H}_{i,v}$. These heatmaps are then embedded into tokens via a patch embedding layer and augmented with learnable positional encodings before being fed into the transformer. The tokens are processed by three successive ViT encoder layers, which effectively model cross-joint and cross-view dependencies. Finally, for the $i$-th keypoint, all tokens corresponding to its multi-view heatmaps (i.e., $\boldsymbol{H}'_{i,L}$ and $\boldsymbol{H}'_{i,R}$) are concatenated and passed through an MLP to regress its 3D position $\boldsymbol{J}_{Cam}^i$ in the camera coordinate system. Meanwhile, the visibility scores from the left and right views are averaged to produce the final 3D visibility score: $\boldsymbol{S}_{Vis} = (\boldsymbol{S}_{L} + \boldsymbol{S}_{R})/2$.

\subsection{Iterative Intra-and Inter-Frame Attention}
\label{subsec:iterative_3d_refinement}
We propose an iterative intra-and inter-frame attention network to further improve both the accuracy and temporal smoothness of pose estimation. First, the 3D keypoints $\boldsymbol{J}_{Cam}^{1:T}$ in the camera coordinate system are transformed into the canonical coordinate system as $\boldsymbol{J}_{Can}^{1:T}$ using the camera poses. The detailed derivation of coordinate transformations is provided in the supplementary material. The canonical 3D keypoints $\boldsymbol{J}_{Can}^{t}$ and visibility scores $\boldsymbol{S}_{Vis}^{t}$ at each frame are then concatenated and processed by a feed-forward network to generate frame-level queries $\{\boldsymbol{q}_{0}^t\}_{t=1}^T$. As shown in Fig.~\ref{fig:overview} (c), these queries undergo alternating processing between a Stereo Transformer Decoder (STD) and a Temporal Transformer Encoder (TTE).
The STD fuses stereo visual features within each frame, while the TTE aggregates data across the entire temporal window. Formally, at the $n$-th iteration, given features $\{\boldsymbol{q}_{n-1}^t\}_{t=1}^T$ from the previous iteration, the STD employs two transformer decoders to enable each $\boldsymbol{q}_{n-1}^t$ to interact with the visual features from the left and right views independently:
\begin{equation}
\boldsymbol{f}_{v}^t=\text{Decoder}(\boldsymbol{q}_{n-1}^t, \boldsymbol{F}_{v}), \quad v\in\{L, R\}
\label{eq:cross_attention}
\end{equation}
The resulting features $\boldsymbol{f}_{L}^t$ and $\boldsymbol{f}_{R}^t$ are concatenated and passed through an MLP to produce a multi-view fused feature $\boldsymbol{f}^t_n$. All multi-view fused features $\{\boldsymbol{f}^t_n\}_{t=1}^T$ within the temporal window are then fed into the TTE, implemented as a transformer encoder, for temporal fusion:
\begin{equation}
[\boldsymbol{q}_{n}^1, \boldsymbol{q}_{n}^2,...,\boldsymbol{q}_{n}^T]=\text{Encoder}([\boldsymbol{f}^1_n, \boldsymbol{f}^2_n,...,\boldsymbol{f}^T_n])
\label{eq:temporal_fusion}
\end{equation}
After $N$ iterations, we obtain multi-view and temporally fused features $\{\boldsymbol{q}_N^t\}_{t=1}^T$, which are subsequently processed by an MLP and a softmax operation to estimate the logits:
\begin{equation}
\bar{\boldsymbol{z}}^{t}=\text{Softmax}(\text{MLP}(\boldsymbol{q}_N^t))
\label{eq:logits_est}
\end{equation}
where $\bar{\boldsymbol{z}}^{t} \in \mathbb{R} ^{M\times K}$ represents the predicted logits for frame $t$. Following prior work~\cite{dwivedi2024tokenhmr}, we avoid the non-differentiable operation of directly selecting an embedding from the codebook by instead estimating logits and multiplying them with the pretrained codebook, yielding approximated quantized features:
\begin{equation}
\boldsymbol{z}^{t}=\bar{\boldsymbol{z}}^{t}_{M\times K} \times \boldsymbol{CB}_{K\times D}
\label{eq:quantized_features_est}
\end{equation}
Finally, the features $\{\boldsymbol{z}^{t}\}_{t=1}^T$ are passed through the VQ-VAE decoder to reconstruct the refined 3D poses $\boldsymbol{J}_{Can}^{1:T}$ in the canonical coordinate system, which are then transformed back to the world coordinate system to produce the final 3D poses $\boldsymbol{J}_{W}^{1:T}$. 

\subsection{Training EvaPose}
\label{subsec:training}
The training process is divided into two stages. 
In the first stage, we introduce a visibility-based loss-weighting scheme to train the visibility-aware 3D pose estimation network. This strategy mitigates interference from invisible keypoints by assigning different loss weights according to their visibility states. The overall loss is defined as: 
\begin{equation}
\mathcal{L}_{\text{stage1}}=\lambda_{\text{vis}} \mathcal{L}_{\text{vis}} + \lambda_{\text{heatmap}} \mathcal{L}_{\text{heatmap}} + \lambda_{\text{3D}} \mathcal{L}_{\text{3D}}
\label{eq:loss_function_stage1}
\end{equation}
where $\lambda_{\text{vis}}$, $\lambda_{\text{heatmap}}$, and $\lambda_{\text{3D}}$ balance the individual loss terms. $\mathcal{L}_{\text{vis}}$ is the binary cross-entropy loss between the predicted and ground-truth visibility labels. The heatmap loss $\mathcal{L}_{\text{heatmap}}$ and the 3D pose loss $\mathcal{L}_{\text{3D}}$ are formulated as:
\begin{equation}
\mathcal{L}_{\text{heatmap}}=\frac{1}{2N_J} \sum_{j=1}^{2}\sum_{i=1}^{N_J} w(s_{i,j}) \cdot  
 \text{MSE}(\boldsymbol{H}_{i,j}, \bar{\boldsymbol{H}}_{i,j}) 
\label{eq:loss_heatmap}
\end{equation}
\begin{equation}
\mathcal{L}_{\text{3D}}=\frac{1}{N_J} \sum_{i=1}^{N_J} \frac{w(s_{i,1}) + w(s_{i,2})}{2} \cdot  
 \text{MSE}(\boldsymbol{J}_{Cam}^i, \bar{\boldsymbol{J}}_{Cam}^i) 
\label{eq:loss_3d_pose_stage1}
\end{equation}
where $\boldsymbol{H}_{i,j}$ and $\bar{\boldsymbol{H}}_{i,j}$ denote the predicted and ground-truth heatmaps for the $i$-th keypoint in the $j$-th camera view. $\boldsymbol{J}_{Cam}^i$ and
$\bar{\boldsymbol{J}}_{Cam}^{i}$ represent the predicted and ground-truth 3D positions of the $i$-th keypoint in the camera coordinate system. The terms $s_{i,1}$ and $s_{i,2}$ indicate the visibility states of the $i$-th keypoint in the left and right image, respectively. The weighting function $w(\cdot)$ assigns a value of 1.0 to visible keypoints and 0.1 to invisible ones, thereby reducing the influence of unreliable invisible keypoints during training.

In the second stage, we train the iterative intra-and inter-frame attention network using a combination of a joint position loss $\mathcal{L}_{\text{joint}}$ and a smooth loss $\mathcal{L}_{\text{smooth}}$. The joint loss $\mathcal{L}_{\text{joint}}$ is computed as the mean squared error between the predicted and ground-truth 3D joint positions. Following~\cite{dai2024hmd}, $\mathcal{L}_{\text{smooth}}$ is defined as the mean absolute error between the predicted and ground-truth joint accelerations.



\section{Experiments}
\label{sec:experiments}
\begin{table*}[tp]
  \centering
  \small
  \begin{tabular}{l l | c c c c c c c c}
    \toprule
    Dataset & Methods & MPJPE$\downarrow$ & PA-MPJPE$\downarrow$ & U-PE$\downarrow$ & L-PE$\downarrow$  & FootPE$\downarrow$  & HandPE$\downarrow$ & Jitter$\downarrow$ &FPS$\uparrow$ \\
    \midrule

    & UnrealEgo~\cite{akada2022unrealego} &60.7 &38.8 &46.5 &81.2 &104.9 &86.6 &7.6 &58.9 \\
    \multirow{3}{*}{\textbf{Eva-3M}}
      & EgoPoseFormer~\cite{yang2024egoposeformer}  &52.9 &33.7 &40.6 &70.5 &93.4 &79.2 &15.4 &\underline{67.0} \\
      & FRAME~\cite{camiletto2025frame} &49.8 &35.1 &42.4 &60.5 &77.4 &80.2 &\underline{3.1} &\textbf{108.7}\\
      & EvaPose-ResNet50 (Ours) &\underline{35.6} &\underline{24.7} &\underline{28.3} &\underline{46.1} &\underline{58.4} &\underline{49.6} &\textbf{3.0} &48.0 \\
      & EvaPose-ViT-L (Ours) &\textbf{34.2} &\textbf{24.0} &\textbf{27.2} &\textbf{44.5} &\textbf{56.4} &\textbf{49.2} &3.2 &9.4\\
    \midrule

    & UnrealEgo~\cite{akada2022unrealego} &55.2 &38.6 &39.9 &77.3 &100.2 &85.2 &5.9 &58.9\\
    \multirow{3}{*}{\textbf{EMHI: P1}}
      & EgoPoseFormer~\cite{yang2024egoposeformer}  &50.7 &29.9 &32.9 &76.5 &96.6 &51.0 &7.3 &\underline{67.0} \\
      & FRAME~\cite{camiletto2025frame} &37.4 &28.2 &\underline{28.0} &51.1 &62.1 &45.1 &\textbf{2.4} &\textbf{108.7}\\
      & EvaPose-ResNet50 (Ours)  &\underline{36.2} &\underline{28.0} &30.0 &\underline{45.3} &\underline{55.7} &\underline{44.8} &\underline{2.6} &48.0 \\
      & EvaPose-ViT-L (Ours) &\textbf{31.7} &\textbf{25.2} &\textbf{24.9} &\textbf{41.9} &\textbf{54.5} &\textbf{39.9} &3.0 &9.4
\\
    \midrule
    
    & UnrealEgo~\cite{akada2022unrealego} &63.7 &42.8 &46.4 &88.7 &115.5 &100.8 &6.1 &58.9\\
    \multirow{3}{*}{\textbf{EMHI: P2}}
      & EgoPoseFormer~\cite{yang2024egoposeformer}  &62.6 &32.7 &42.0 &92.3 &118.1 &66.8 &7.6 &\underline{67.0} \\
      & FRAME~\cite{camiletto2025frame} &60.5 &44.3 &55.8 &67.4 &78.6 &75.5 &6.4 &\textbf{108.7}\\
      & EvaPose-ResNet50 (Ours) &\underline{38.5} &\underline{29.5} &\underline{31.4} &\underline{48.8} &\underline{61.9} &\underline{57.8} &\textbf{3.1} &48.0 \\
      & EvaPose-ViT-L (Ours) &\textbf{33.3} &\textbf{26.2} &\textbf{25.4} &\textbf{44.7} &\textbf{58.9} &\textbf{47.5} &\underline{3.4} &9.4\\
    \bottomrule
  \end{tabular}
  \caption{Comparison with state-of-the-art methods on the Eva-3M and EMHI datasets. The best and second best results are highlighted in \textbf{boldface} and \underline{underlined}, respectively.}
  \label{tab:result_withsota_methods}
\end{table*}

\begin{table*}[htp]
  \centering
  \small
  \begin{tabular}{l | c c | c c | c c | c c | c c}
    \toprule
    \multirow{2}{*}{Methods} 
    & \multicolumn{2}{c|}{Ankle} & \multicolumn{2}{c|}{Foot} & \multicolumn{2}{c|}{Elbow} & \multicolumn{2}{c|}{Wrist} & \multicolumn{2}{c}{Mean} \\
    & Visible & Invisible & Visible & Invisible & Visible & Invisible & Visible & Invisible & Visible & Invisible \\
    \midrule
    UnrealEgo~\cite{akada2022unrealego} &83.0 &87.1 &91.3 &106.8 &52.0 &63.0 &67.0 &85.8 &73.3 &85.7\\
    EgoPoseFormer~\cite{yang2024egoposeformer} &78.2 &94.2 &83.9 &96.4 &49.1 &63.2 &57.9 &101.0 &67.3 &88.7\\
    FRAME~\cite{camiletto2025frame} &77.8 &79.1 &81.2 &92.5 &53.0 &59.5 &67.9 &86.5 &70.0 &79.4\\
    EvaPose-ResNet50 &\underline{52.2} &\underline{66.4} &\underline{60.8} &\underline{82.5} &\underline{32.2} &\underline{48.0} &\underline{38.0} &\underline{65.7} &\underline{45.8} &\underline{65.6}\\
    EvaPose-ViT-L &\textbf{49.2} &\textbf{63.8} &\textbf{57.2} &\textbf{79.4} &\textbf{29.5} &\textbf{45.6} &\textbf{34.2} &\textbf{63.3} &\textbf{42.5} &\textbf{63.0}\\
    \bottomrule
  \end{tabular}
  \caption{Performance comparisons of visible and invisible keypoints on the Eva-3M dataset.}
  \label{tab:results_vis_invis_analysis}
\end{table*}

\begin{figure*}[htp]
  \centering
  \includegraphics[width=0.95\linewidth]{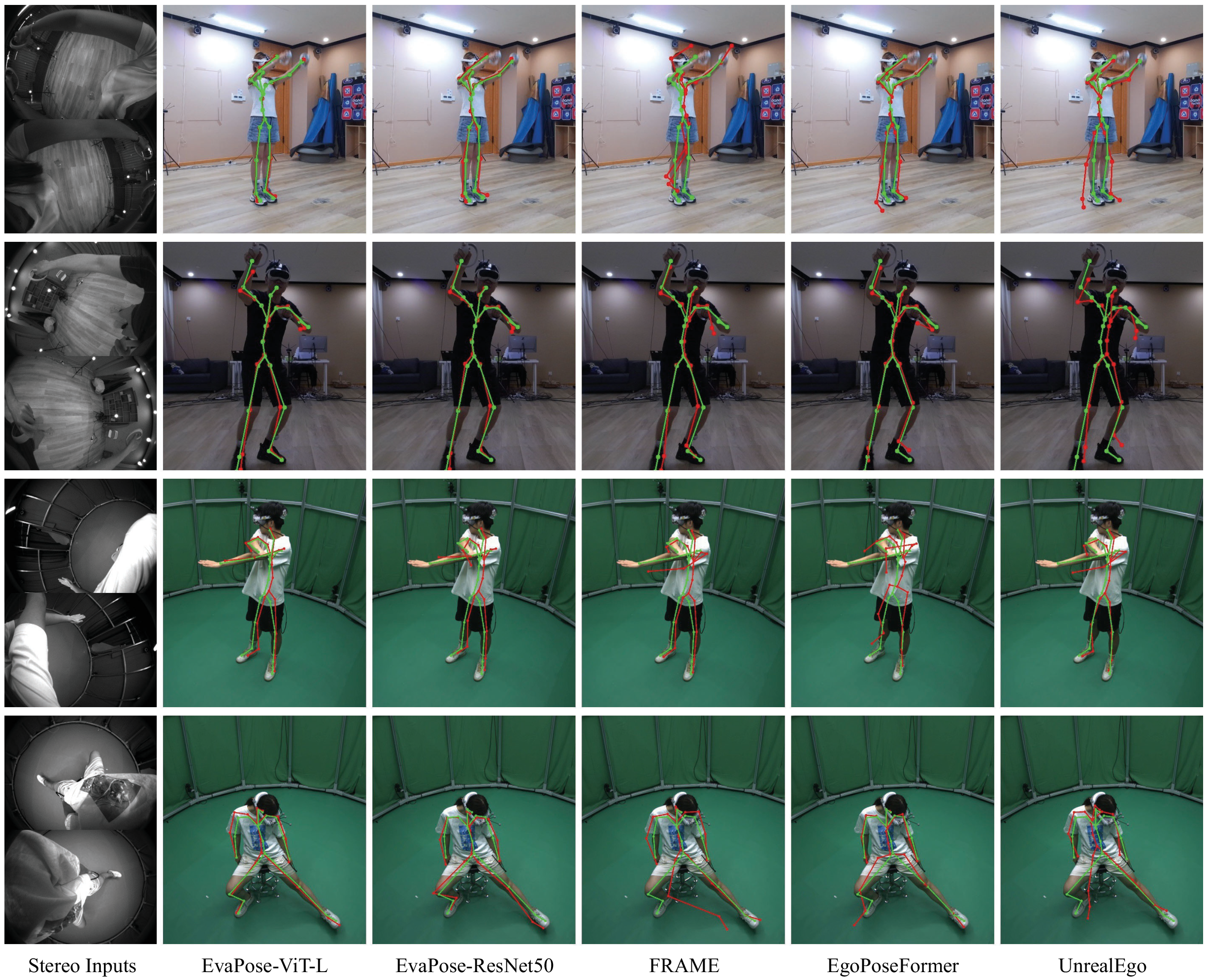}
  \caption{Qualitative comparisons on both EMHI (the first two rows) and Eva-3M (the last two rows) datasets. For better illustration, the predicted (red) and ground-truth (green) 3D poses are re-projected onto external reference views which are not used for pose estimation.}
  \label{fig:qualitative_comparison}
\end{figure*}

\begin{table}[t]
  \centering
  \footnotesize
  \begin{tabular}{c|cccc}
    \toprule
    Method & MPJPE$\downarrow$ & PA-MPJPE$\downarrow$ & VLK-PE$\downarrow$  & ILK-PE$\downarrow$ \\
    \midrule
    w/o visibility  &40.6 &27.9 &53.5 &\textbf{65.3}\\
    with visibility &\textbf{35.6} &\textbf{24.7} &\textbf{45.8} &65.6\\
    \bottomrule
  \end{tabular}
  \caption{Impact of the visibility-aware model and loss design. VLK-PE is the MPJPE for visible limb keypoints, including ankle, foot, elbow and wrist. ILK-PE denotes the MPJPE for invisible limb keypoints.}
  \label{tab:ablation_of_visibility}
\end{table}

\begin{table}[t]
  \centering
  \footnotesize
  \begin{tabular}{ccc|ccc}
    \toprule
    TTE & STD & VQ-VAE & MPJPE$\downarrow$ & PA-MPJPE$\downarrow$ & Jitter$\downarrow$\\
    \midrule
    \ding{55} & \ding{55} & \ding{55} &46.0 &33.3 &4.7 \\
    \ding{51} & \ding{55} & \ding{55} &40.3 &27.0 &\textbf{2.5}\\
    \ding{51} & \ding{51} & \ding{55} &39.1 &25.9 &2.7\\
    \ding{51} & \ding{51} & \ding{51} &\textbf{35.6} &\textbf{24.7} &3.0 \\
    \bottomrule
  \end{tabular}
  \caption{Ablation study for key components in our model.}
  \label{tab:ablation_of_keycomponents}
\end{table}

\subsection{Dataset}
\noindent
\textbf{Eva-3M.}
Our dataset comprises 1,353 motion sequences from 31 subjects performing 24 categories of actions common in VR scenarios. We reserve all sequences from two randomly chosen subjects (one male and one female, 98 sequences total) for testing, using the remainder for training.

\noindent
\textbf{EMHI~\cite{fan2025emhi}.}
We adhere to the official split, dividing the data into three subsets: 70$\%$ for training, and two separate test sets, i.e., P1 (16$\%$) and P2 (14$\%$). Notably, P2 contains unseen action categories and is designed specifically to evaluate the model’s generalization capability to out-of-distribution motions. We further annotate 488K frames in this dataset with keypoint visibility labels to support visibility-aware model training and evaluation.


\subsection{Implementation Details and Metrics}
\noindent
\textbf{Implementation Details.}
For the VQ-VAE module, we set the number of tokens to $M$ = 160, and the codebook size to 2048 $\times$ 256. The VQ-VAE network is pre-trained on the AMASS~\cite{mahmood2019amass}, MOYO~\cite{tripathi20233d} and AIST++~\cite{li2021ai} datasets, and remains frozen during the training of EvaPose. For EvaPose, we employ two image backbones: ResNet50~\cite{he2016deep} and ViT-Large~\cite{dosovitskiy2020vit, oquab2023dinov2}, with the corresponding models denoted as EvaPose-ResNet50 and EvaPose-ViT-L. Input images are resized to 448$\times$336 for EvaPose-ViT-L, and kept at the original 640$\times$480 resolution for EvaPose-ResNet50. Training proceeds in two stages. First, the visibility-aware 3D pose estimation module is trained with a batch size of 24 and learning rate of $1\times10^{-5}$ for 20 epochs. The weights are set as $\lambda_{\text{vis}}$ = $5\times10^{-3}$, $\lambda_{\text{heatmap}}$ = 0.1, and $\lambda_{\text{3D}}$ = 1.0. Then, the temporal 3D pose refinement network is trained with a batch size of 4 and a learning rate of $1\times10^{-5}$ for 40 epochs, using a time window of $T$ = 24 frames. Further architectural details are provided in the supplementary material.


\noindent
\textbf{Metrics.}
As in prior works, we evaluate pose estimation accuracy using six metrics: mean per joint positional error (MPJPE), Procrustes-aligned MPJPE (PA-MPJPE), upper-body MPJPE (U-PE), lower-body MPJPE (L-PE), foot MPJPE (FootPE), and hand MPJPE (HandPE). All values are reported in millimeters. To assess motion smoothness, we adopt Jitter metric ($10^2m/s^3$) as in~\cite{dai2024hmd}. Inference speed is measured in frames per second (FPS), evaluated on a single NVIDIA V100 GPU under consistent settings.


\subsection{Comparison with State-of-the-Art Methods}
To ensure a rigorous comparison, we retrain three leading egocentric HPE methods, i.e., UnrealEgo~\cite{akada2022unrealego}, EgoPoseFormer~\cite{yang2024egoposeformer}, and FRAME~\cite{camiletto2025frame}, on both the EMHI and our Eva-3M datasets, adhering to their original experimental settings. As summarized in Tab.~\ref{tab:result_withsota_methods}, our
approach consistently outperforms prior methods across most key metrics on all test sets. Notably, on the EMHI P2 test set, which contains actions unseen during training, both EvaPose-ResNet50 and EvaPose-ViT-L achieve substantial improvements over the previous state-of-the-art, FRAME~\cite{camiletto2025frame}, in both MPJPE and Jitter metrics. These experimental results validate the superiority of our method not only in terms of accuracy and smoothness of pose estimation, but also in generalization capability to unseen actions when applying to real VR scenarios.


To further understand how EvaPose-ResNet50 and EvaPose-ViT-L outperform previous methods, we compare the detailed accuracy on visible and invisible keypoints using the Eva-3M dataset.  Tab.~\ref{tab:results_vis_invis_analysis} presents a breakdown for representative limb keypoints, from which we draw the following observations: (1) Previous methods yield comparable accuracy on visible limb keypoints, yet perform considerably worse than our approach. This suggests that indiscriminately treating visible and invisible keypoints degrades their estimation accuracy on visible ones. (2) Compared to FRAME~\cite{camiletto2025frame}, our EvaPose-ResNet50 and EvaPose-ViT-L bring more pronounced accuracy improvements on visible keypoints than on invisible ones, indicating that our EvaPose can effectively utilize visibility information to mitigate interference from invisible joints. (3) Ground-truth visibility annotations are highly valuable in egocentric HPE, where self-occlusion and out-of-FoV are common.


In terms of inference speed, EvaPose-ResNet50 runs at 48.0 FPS on an NVIDIA V100 GPU, enabling real-time performance despite being slower than prior methods. EvaPose-ViT-L, which employs a large ViT-Large backbone with 0.3B parameters, achieves the highest accuracy across all datasets at the cost of lower FPS, as expected.

The qualitative results in Fig.~\ref{fig:qualitative_comparison} further highlight our method's superior capability in estimating accurate and physically plausible 3D poses, including those of invisible keypoints. A significant enhancement is evident in the limb regions, which is consistent with the quantitative results in Tab.~\ref{tab:result_withsota_methods}. We also provide more quantitative and qualitative comparison results in our supplementary materials. 

\subsection{Ablation Study}
We conduct comprehensive ablation studies on the Eva-3M dataset using the ResNet50 image backbone to evaluate the contribution of each proposed module.

\noindent
\textbf{Impact of the visibility-aware model and loss design.} 
As one of our key contributions, we introduce a visibility-aware 3D pose estimation network along with a visibility-based loss-weighting scheme to explicitly model keypoint visibility and leverage it for improved pose estimation accuracy. To validate their effectiveness, we perform an ablation by removing all visibility-related components. As reported in Tab.~\ref{tab:ablation_of_visibility}, incorporating the visibility-aware model and loss design significantly reduces MPJPE for visible limb keypoints, leading to notable improvements in both overall MPJPE and PA-MPJPE metrics. These results jointly demonstrate the importance of visibility annotations and the efficacy of our visibility-aware modeling approach.

\noindent
\textbf{Impact of the key components.}
As described in Section~\ref{sec:method}, there are three key components in our model: the VQ-VAE pose prior, the Stereo Transformer Decoder (STD), and the Temporal Transformer Encoder (TTE). Tab.~\ref{tab:ablation_of_keycomponents} summarizes our ablation study by removing these components sequentially to isolate their contributions. First, introducing the pose prior via VA-VAE decreases MPJPE from 39.1mm to 35.6mm, demonstrating that 3D pose priors learned from large-scale mocap datasets can help resolve ambiguities, particularly for invisible keypoints. Second, incorporating stereo visual features via the STD network can contribute to additional accuracy improvement. Third, the TTE module plays an important role in reducing both MPJPE and Jitter metrics, which confirms the importance of temporal fusion in improving smoothness and accuracy.

\section{Conclusion}
\label{sec:conclusion}
We introduced Eva-3M, a large-scale real egocentric visibility-aware HPE dataset. It provides  visibility labels for the first time and surpasses existing datasets in motion diversity. Captured using a Pico4 Ultra VR-MR device, Eva-3M offers realistic insights into HPE in real VR-MR devices. We also proposed EvaPose, a novel egocentric visibility-aware HPE method explicitly leveraging the visibility information to improve the HPE accuracy. Extensive experiments validate that EvaPose achieves state-of-the-art performance on both the EMHI and Eva-3M benchmarks.

\noindent
\textbf{Limitations.} 
As a data-driven approach, our method relies on large-scale, high-quality GT labels, which are challenging to acquire in in-the-wild scenarios. Exploring weakly-supervised or self-supervised training methods to improve generalization will be a promising direction for future work.

{
    \small
    \bibliographystyle{ieeenat_fullname}
    \bibliography{main}
}

\clearpage
\setcounter{page}{1}
\maketitlesupplementary

\section{Dataset Acquisition}
\textbf{Capture Setup.} Our capture rig consists of a Pico4 Ultra VR-MR HMD device, 16 industrial cameras with image resolution of 2488$\times$2048 and 10 OptiTrack cameras. All recordings are conducted within a cylindrical capture volume of 4m in diameter. We temporally synchronize the egocentric cameras and multi-view recordings via the OptiTrack system, and spatially align them into a unified coordinate frame. Ground-truth poses are then obtained by multi-view triangulation followed by SMPL fitting.

\noindent\textbf{Temporal Synchronization.} The industrial cameras and OptiTrack system are hardware‐synchronized using a dedicated sync hub device, ensuring that all external cameras and motion capture hardware share a common trigger signal. Meanwhile, the Pico4 Ultra VR-MR HMD is synchronized offline. Specifically, an optical rigid body (ORBs) marker is mounted to the VR device, and we align its motion with the OptiTrack data using a motion‐correlation algorithm~\cite{9271875}. Finally, both egocentric camera frames and external recordings are mapped into the same timeline via OptiTrack. Since the egocentric images are captured at 30Hz, a maximum temporal misalignment of approximately 16.5ms may persist. To mitigate this problem, the final SMPL parameters are post‐processed using interpolation to better align with the egocentric images.

\noindent\textbf{Spatial Alignment.} We attach optical rigid body markers to the HMD device in order to capture HMD's 6-DoF pose \(\boldsymbol{T}_{head}^{opti}\) in the OptiTrack coordinate system. Formally,
\[
\boldsymbol{T}_{head}^{opti} = \boldsymbol{T}_{orb}^{opti} \cdot \boldsymbol{T}_{head}^{orb},
\]
where \(\boldsymbol{T}_{orb}^{opti}\) is the pose of the ORB in OptiTrack, and \(\boldsymbol{T}_{head}^{orb}\) is the rigid transform from HMD to ORB which can be pre-calibrated. We apply the same method to compute the pose of each industrial camera. Thus we obtain all necessary transform matrices among the HMD, cameras, and the OptiTrack coordinate systems.

\noindent\textbf{3D Keypoints Estimation.} We first employ the HRNet~\cite{sun2019deep} to extract 2D keypoints in the Body25~\cite{cao2017realtime} format for all industrial camera views. Then, the multi-view 2D detections are used to generate 3D keypoints via an iterative triangulation procedure. Note that the low-confidence detections and outliers with large re-projection errors are filtered out in the iterative triangulation process. To further improve temporal smoothness and physical plausibility, the resulting 3D keypoints $\boldsymbol{p}^{tri} \in \mathbb{R}^{75}$ are optimized with a smoothness regularization and a bone-length constraint, yielding stable and anatomically plausible 3D trajectories.

\noindent\textbf{SMPL Fitting.} We perform multi‐stage optimization to fit SMPL parameters $\theta \in \mathbb{R}^{75}$ and $\beta \in \mathbb{R}^{10}$ to the triangulated keypoints $\boldsymbol{p}^{tri}$. In the first stage, we optimize the body shape parameters $\beta \in \mathbb{R}^{10}$ by minimizing:  
\[
\mathcal{L}_{shape}(\beta) = \sum_{(i,j)\in \mathcal{B}} \|\, \|\boldsymbol{v}_i - \boldsymbol{v}_j\| - l_{ij}\|^2 + \|\beta\|^2
\]  
where \(\mathcal{B}\) is a set of bone edges, \(\boldsymbol{v}_i, \boldsymbol{v}_j\) are the SMPL joint positions in rest pose, and \(l_{ij}\) is the target bone length from the triangulated keypoints. The second term serves as a regularization to penalize large shape values. 

\begin{figure*}[htp]
  \centering
  \includegraphics[width=0.95\linewidth]{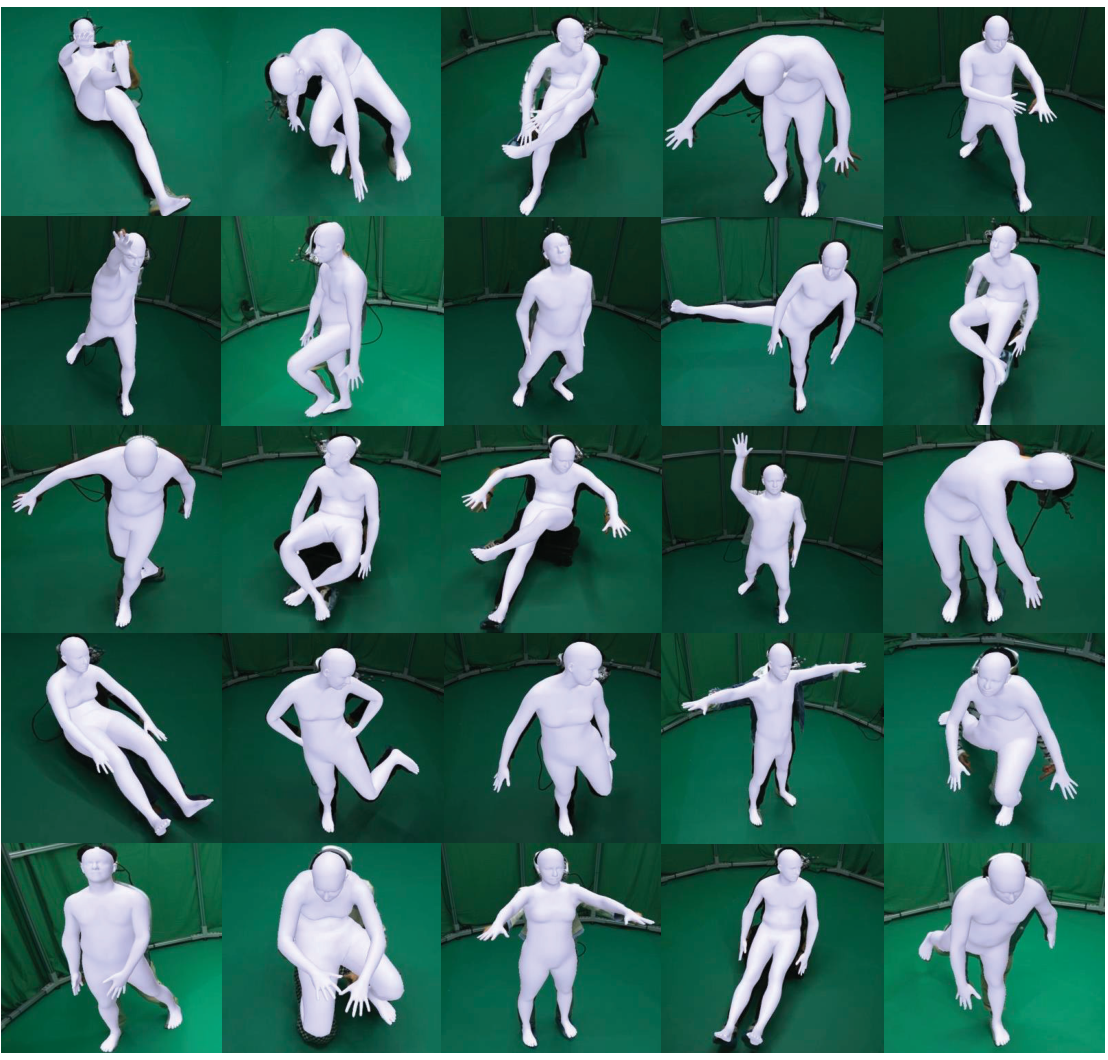}
  \caption{Visualization of some representative fitted SMPL meshes.}
  \label{fig:gt_smpl}
\end{figure*}

Next, we optimize the global rotation \(R\) and translation \(T\), and finally the local pose parameters by minimizing the following loss function:
\begin{equation}
\begin{split}
\mathcal{L}_{pose}(\beta, \theta) = \; &
\lambda_{fk} \mathcal{L}_{fk} +
\lambda_{smooth} \mathcal{L}_{smooth} \\ 
&+ \lambda_{reg} \mathcal{L}_{reg} +
\lambda_{init} \mathcal{L}_{init}
\end{split}
\end{equation}
where $\lambda_{fk}$, $\lambda_{smooth}$, $\lambda_{reg}$, and $\lambda_{init}$ are the weights for the respective loss terms. $\mathcal{L}_{fk}$ calculates the mean square errors between the triangulated keypoints and the SMPL forward-kinematics (FK) keypoints. Note that we use the regression matrix of the Body25 format in this FK process to adapt to the triangulated keypoints $\boldsymbol{p}^{tri}$. $\mathcal{L}_{smooth}$ is a temporal smoothness term, $\mathcal{L}_{reg}$ is a regularization term which penalizes large pose parameters. Since optimization is performed in batches of 10 frames, the first frame of each batch is initialized from the last frame of the previous one, and an additional loss $\mathcal{L}_{init}$ is applied to prevent the solution from drifting too far. In Fig.~\ref{fig:gt_smpl}, we visualize some representative fitted SMPL meshes that are re-projected onto external industrial cameras for better illustration.

After obtaining SMPL parameters $\theta \in \mathbb{R}^{75}$ and $\beta \in \mathbb{R}^{10}$ as described above, we can generate the 3D keypoints in the SMPL format via a standard FK process. And these 3D keypoints are used for model training and evaluation.

\begin{figure*}[htp]
  \centering
  \includegraphics[width=1.0\linewidth]{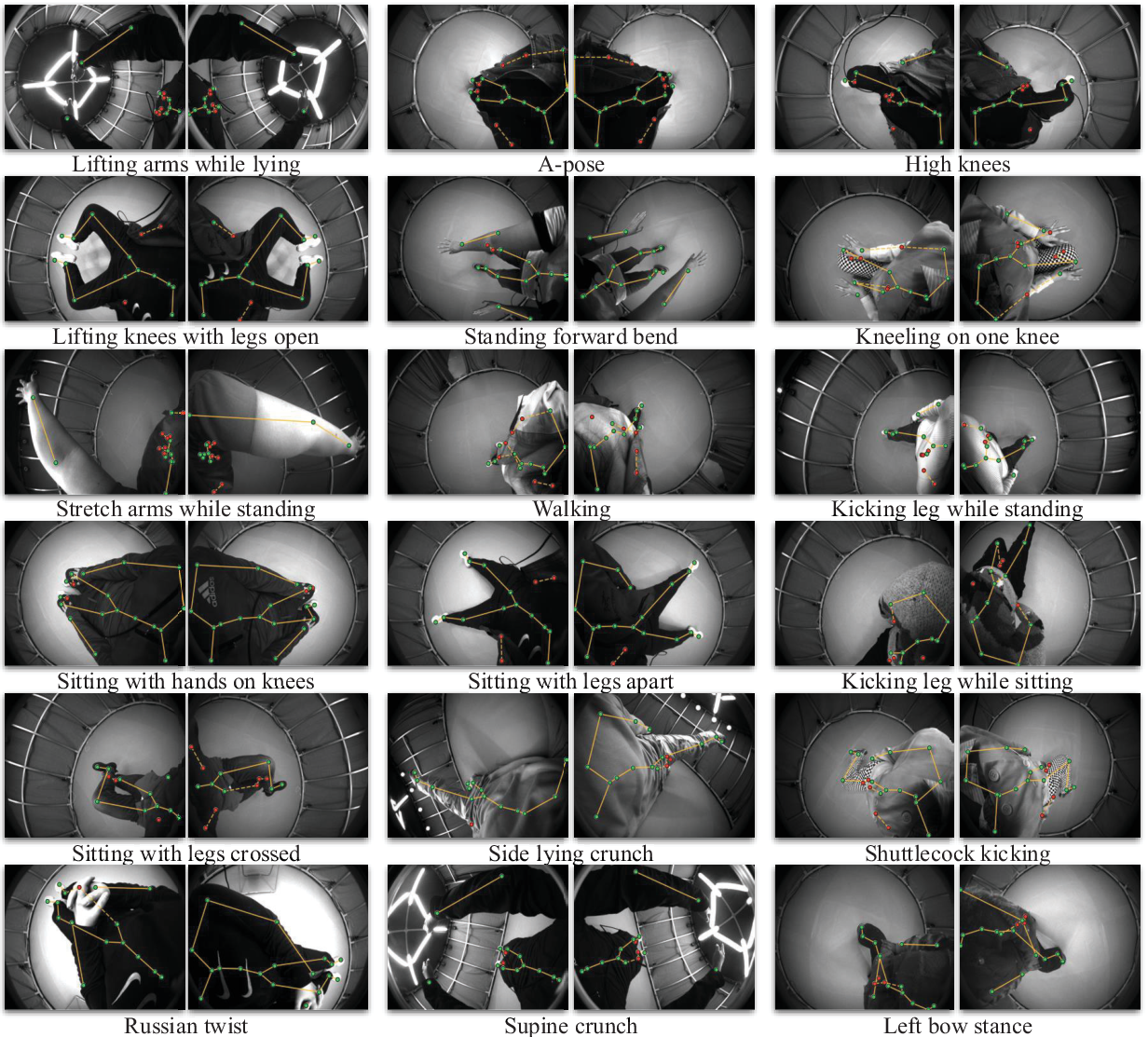}
  \caption{Representative samples of the keypoint visibility labels in our Eva-3M dataset.}
  \label{fig:gt_visibility}
\end{figure*}

\noindent\textbf{Visibility Labeling.} 
Recall that we have collected 3.0M frames of stereo egocentric image pairs in our Eva-3M dataset. To improve the keypoint visibility labeling efficiency, we discard the first 600 calibration frames of each sequence and annotate keypoint visibility every 10 frames thereafter. Specifically, we project SMPL keypoints onto the stereo egocentric images and require professional annotators to manually annotate the visibility of each keypoint on each image. In total, we annotate 435K frames with keypoint visibility labels. In Fig.~\ref{fig:gt_visibility}, we presents a few representative samples of the keypoint visibility labels in our Eva-3M dataset. It demonstrates that the keypoint invisibility issue is evident in the egocentric settings due to self-occlusions and out-of-FoV.


\begin{table*}[htp]
  \centering
  \small
  \begin{tabular}{l l c| c c c}
    \toprule
    Dataset & Methods &Backbone & MPJPE$\downarrow$ & PA-MPJPE$\downarrow$ & Jitter$\downarrow$ \\
    \midrule

    & UnrealEgo~\cite{akada2022unrealego} & ResNet18 &289.7 &171.0 & 60.4 \\
    \multirow{3}{*}{\textbf{EMHI: P1}}
      & EgoPoseFormer~\cite{yang2024egoposeformer}  & ResNet18 &180.5 &112.5 &70.3 \\
      & FRAME~\cite{camiletto2025frame} & ResNet50 &135.6 &126.7 &12.9 \\
      & EvaPose-ResNet50 (Ours) & ResNet50 &93.3 & 59.5 &\textbf{5.0} \\
      & EvaPose-ViT-L (Ours) & ViT-Large &\textbf{49.5} &\textbf{38.4} &7.6 \\
    \midrule
    
    & UnrealEgo~\cite{akada2022unrealego} & ResNet18 &244.9 &148.5 &40.4 \\
    \multirow{3}{*}{\textbf{EMHI: P2}}
      & EgoPoseFormer~\cite{yang2024egoposeformer}  & ResNet18 &160.7 &107.3 &60.2\\
      & FRAME~\cite{camiletto2025frame} & ResNet50 & 112.7 &106.9 &9.8\\
      & EvaPose-ResNet50 (Ours) & ResNet50 & 103.6 &57.4 &\textbf{4.7}\\
      & EvaPose-ViT-L (Ours) & ViT-Large &\textbf{45.9} &\textbf{36.0} &6.6\\
    \bottomrule
  \end{tabular}
  \caption{Cross-dataset evaluation results. In this experiment, all methods are trained on the Eva-3M dataset and tested on the EMHI dataset.}
  \label{tab:result_cross_dataset}
\end{table*}

\section{Implementation Details}
In this section, we provide additional implementation details of our EvaPose model.

\subsection{Visibility-Aware 3D Pose Estimation}
\noindent
\textbf{Image Encoder.} Recall that our EvaPose uses two image backbones to extract visual features: ResNet50~\cite{he2016deep} and ViT-Large~\cite{dosovitskiy2020vit}. For the ResNet50 image encoder, following previous works~\cite{camiletto2025frame}, we use the weights pretrained on ImageNet~\cite{deng2009imagenet} dataset. For the ViT-Large image encoder, we use the pretrained ViT-L/14 weights from DINOv2~\cite{oquab2023dinov2} to benefit from the self-supervised pretraining on large quantities of data. To lower the computational load, the input images are resized to 448$\times$336 for the ViT-Large backbone.

\noindent
\textbf{Heatmap Estimation.} Motivated by ViTPose~\cite{xu2022vitpose}, we use a  lightweight decoder to process the features extracted from the backbone network and localize the keypoints. It consists of three stacks of deconvolution blocks and one 2D convolution layer. Each deconvolution block contains a 2D transposed convolution layer which upsamples the image feature maps by 2 times, followed by batch normalization~\cite{ioffe2015batch} and ReLU~\cite{agarap2018deep} activation. The convolution layer with the kernel size 1$\times$1 is utilized to get the heatmaps for the keypoints.

\noindent
\textbf{Visibility Prediction.} For simplicity, we use a convolution layer and a MLP network to capture the visibility status of individual keypoints. First, the image feature maps are fed into a 2D convolution layer, reducing its channel dimension to the number of keypoints. Then, the 2D image feature maps are flattened and passed through a three-layer MLP network to predict a visibility score for each keypoint.

\noindent
\textbf{Estimating 3D Poses in the Camera Coordinate System.} Given the multi-view visibility-aware heatmaps as input, we first embed the heatmaps into tokens via a patch embedding layer. Then, a set of learnable positional encodings is added to the tokens,  resulting in the transformer input. After that, the embedded tokens are processed by three transformer layers, each of which is consisted of a multi-head self-attention (MHSA) layer and a feed-forward network (FFN). Each transformer layer has a feedforward dimension of 512 and 8 attention heads. Finally, the output tokens corresponding to individual keypoint heatmaps are concatenated and projected through a three-layer MLP network to output the 3D positions in the camera coordinate system.

\subsection{Iterative Intra-and Inter-Frame Attention}
\noindent
\textbf{Stereo Transformer Decoder.} 
The Stereo Transformer Decoder (STD) network is used for multi-view feature fusion within each frame. First, it applies two transformer decoders to let the query feature interact with the image features of each viewpoint independently. Each transformer decoder consists of 4 transformer decoder layers, each with a feedforward dimension of 512 and 8 attention heads. Then, the multi-view features are concatenated and fed into a one-layer MLP network for fusion.

\noindent
\textbf{Temporal Transformer Encoder.} Having  the estimated per-frame multi-view fused features, we use a Temporal Transformer Encoder (TTE) for temporal fusion of all information within a time window of $T=24$. Specifically, TTE  is structured with 8 transformer encoder layers, each having a feedforward dimension of 512 and 8 attention heads.

\begin{figure}[tp]
  \centering
  \includegraphics[width=0.6\linewidth]{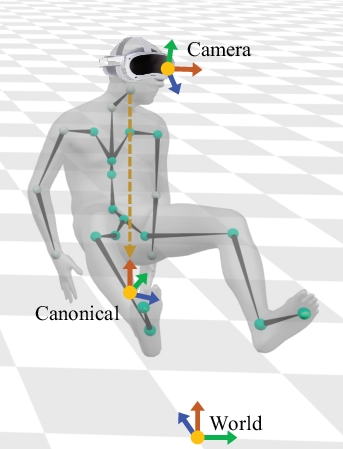}
  \caption{Overview of the coordinate systems.}
  \label{fig:coordinate_systems}
\end{figure}

\section{Coordinate Systems Transformation}
As shown in Fig.~\ref{fig:coordinate_systems}, there are three coordinate systems used in this paper: the camera coordinate system which is defined as the left camera coordinate system, the world coordinate system, and the canonical coordinate system. In this section, we provide the relative transformations among these coordinate systems.

\noindent
\textbf{Camera-to-World Transformation.} As described in the main paper, the SLAM system of the Pico4 Ultra VR-MR HMD device can provide the camera poses in the world coordinate system at each timestamp $t$, including camera rotation $\boldsymbol{R}_{world, cam}^t$ and position $\boldsymbol{T}_{world, cam}^t$ with millimeter-level accuracy. Formally, the camera-to-world transformation matrix can be computed as:
\begin{equation}
\boldsymbol{M}^t_{world,cam}=(\boldsymbol{R}_{world, cam}^t, \boldsymbol{T}_{world, cam}^t) \in \text{SE}(3)
\label{eq:camera2world}
\end{equation}

\noindent
\textbf{Canonical-to-World Transformation.} As illustrated in Fig.~\ref{fig:coordinate_systems},
the canonical coordinate system is obtained by projecting the head joint to the floor plane, aligning $y$ axis to be vertical, and using the projection of the horizontal axis of head orientation on the floor plane as the horizontal axis. Formally, given the estimated head joint position in the world coordinate system $\boldsymbol{T}_{world, head}^t$, we can compute the canonical-to-world translation matrix:
\begin{equation}
\boldsymbol{T}_{world, can}^t=[\boldsymbol{e}_x, \vec{0} , \boldsymbol{e}_z]^T\boldsymbol{T}_{world, head}^t
\label{eq:Tcan2world}
\end{equation}
For rotation matrix, we align the canonical frame’s $z$-axis toward the "forward" direction of the head pose. 
\begin{equation}
\vec{v}^t=\boldsymbol{R}_{world, head}^t \boldsymbol{e}_y
\label{eq:Rcan2world1}
\end{equation}
\begin{equation}
\boldsymbol{R}_{world, can}^t=\boldsymbol{R}_{y}(-arctan2(\boldsymbol{e}_x^T \vec{v}^t, \boldsymbol{e}_z^T \vec{v}^t))
\label{eq:Rcan2world2}
\end{equation}
where $\boldsymbol{R}_{y}: \mathbb{R} \to \text{SO}(3)$ constructs a y-axis rotation. Then, the canonical-to-world transformation matrix can be computed as:
\begin{equation}
\boldsymbol{M}^t_{world,can}=(\boldsymbol{R}_{world, can}^t, \boldsymbol{T}_{world, can}^t) \in \text{SE}(3)
\label{eq:canonical2world}
\end{equation}

\section{Cross-Dataset Performance Evaluation}
In egocentric human pose estimation, cross-dataset generalization remains a problem. Given that EMHI and Eva-3M were captured using PICO4 and PICO4 Ultra HMD devices, respectively, these two datasets have similar camera settings. The Eva-3M dataset is collected in a controlled laboratory environment with green screen backgrounds. Differently, the EMHI dataset is captured in real indoor environments. To simulate real-world usage, we conduct this cross-generalization experiment using the Eva-3M dataset as the training set and the EMHI dataset as the test set. Tab.~\ref{tab:result_cross_dataset} shows the results. It can be concluded from these results that: (1) Using a larger backbone network can achieve better generalization capabilities. (2) Compared to FRAME~\cite{camiletto2025frame}, our EvaPose-ResNet50 uses the same ResNet50 backbone and achieves better results. This demonstrates that our approach has better generalization performance. (3) Our EvaPose-ViT-L significantly outperforms all other methods in both MPJPE and PA-MPJPE metrics. It shows that the ViT-Large backbone from DINOv2~\cite{oquab2023dinov2}, pretrained on large quantities of data, can improve the generalization performance of the model.

\end{document}